\documentclass[conference]{IEEEtran}
\IEEEoverridecommandlockouts

\usepackage{cite}
\usepackage{amsmath,amssymb,amsfonts}
\usepackage{algorithmic}
\usepackage{graphicx}
\usepackage{textcomp}
\usepackage{xcolor}
\def\BibTeX{{\rm B\kern-.05em{\sc i\kern-.025em b}\kern-.08em
    T\kern-.1667em\lower.7ex\hbox{E}\kern-.125emX}}
\begin{document}

\title{HDiffTG: A Lightweight Hybrid Diffusion-Transformer-GCN Architecture for 3D Human Pose Estimation}

\author{\IEEEauthorblockN{Yajie Fu$^{1}$, Chaorui Huang$^{1}$, Junwei Li$^{1\ast}$, Hui Kong$^{2}$, Yibin Tian$^{3}$, Huakang Li$^{4}$, Zhiyuan Zhang$^{5\ast}$\thanks{${\ast}$ Corresponding author: lijunwei7788@zju.edu.cn, cszyzhang@gmail.com}
}

\IEEEauthorblockA{
$^1$\textit{College of Information Science and Electronic Engineering, Zhejiang University, China}\\
$^2$\textit{Faculty of Science and Technology, University of Macau, China}\\
$^3$\textit{College of Mechatronics and Control Engineering, Shenzhen University, China}\\
$^4$\textit{School of AI and Advanced Computing, Xi'an Jiaotong-Liverpool University, China}\\
$^5$\textit{School of Computing and Information Systems, Singapore Management University, Singapore}
}
}

\maketitle

\begin{abstract}
We propose HDiffTG, a novel 3D Human Pose Estimation (3DHPE) method that integrates Transformer, Graph Convolutional Network (GCN), and diffusion model into a unified framework. HDiffTG leverages the strengths of these techniques to significantly improve pose estimation accuracy and robustness while maintaining a lightweight design. The Transformer captures global spatiotemporal dependencies, the GCN models local skeletal structures, and the diffusion model provides step-by-step optimization for fine-tuning, achieving a complementary balance between global and local features. This integration enhances the model's ability to handle pose estimation under occlusions and in complex scenarios. Furthermore, we introduce lightweight optimizations to the integrated model and refine the objective function design to reduce computational overhead without compromising performance. Evaluation results on the Human3.6M and MPI-INF-3DHP datasets demonstrate that HDiffTG achieves state-of-the-art (SOTA) performance on the MPI-INF-3DHP dataset while excelling in both accuracy and computational efficiency. Additionally, the model exhibits exceptional robustness in noisy and occluded environments. Source codes and models are available at https://github.com/CirceJie/HDiffTG
\end{abstract}


\section{Introduction}
3D Human Pose Estimation (3DHPE) from monocular visual data is a critical task in computer vision, with numerous applications in areas such as human-computer interaction~\cite{chen2019holistic++}, augmented/virtual reality~\cite{mehta2017vnect}, action recognition~\cite{rajasegaran2023benefits}, and motion capture~\cite{desmarais2021review}. Given the wide range of applications, there is an increasing demand for more accurate and computationally efficient pose estimation models.
With significant advancements in 2D pose estimation, the typical solution for 3DHPE today involves breaking down the problem into two sequential stages. A 2D pose detector identifies 2D keypoints using a pre-trained model which are subsequently transformed into 3D joint coordinates in the next stage via a 2D-to-3D pose uplifting algorithm~\cite{drover2018can,wandt2019repnet,einfalt2023uplift}. Despite the higher spatial precision achieved, lifting from 2D to 3D poses additional challenges, as this procedure is highly ill-posed due to depth ambiguities, occluded body parts, and the complexity of human body dynamics~\cite{cheng2019occlusion, yuan2021simpoe, holmquist2023diffpose}.

To overcome these challenges, two types of approach have emerged in recent years: probabilistic and deterministic approaches. Probabilistic methods~\cite{li2019generating,wehrbein2021probabilistic,shan2023diffusion} model the 2D-to-3D lifting as a probability distribution and generate multiple potential solutions for each image, accommodating uncertainty and ambiguity in the lifting process. Although these methods achieve promising results~\cite{MHFormer,holmquist2023diffpose,kim2024mhcanonnet}, performance can degrade when many generated hypotheses deviate from the actual pose. This issue is particularly evident in real-life scenarios where noise and occlusion are common. Generating numerous hypotheses to cover the actual pose often leads to averaged predictions that are inaccurate. Furthermore, generating multiple predictions reduces inference efficiency, which is one of the key objectives our work aims to address.

\begin{figure}[t]
    \centering
    \includegraphics[width=1\linewidth]{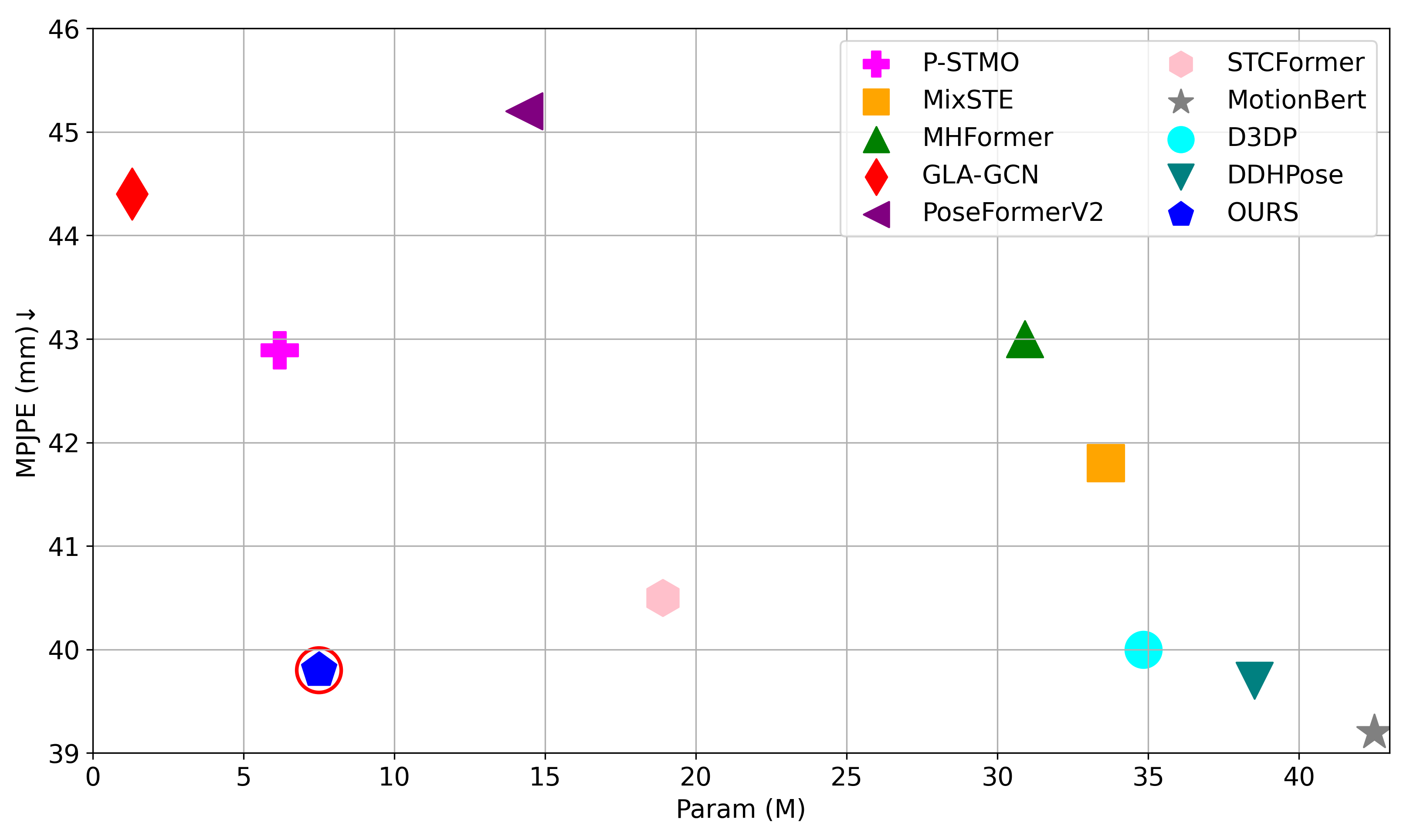}
    \caption{Comparison of different 3D human pose estimation methods on the Human3.6M dataset in terms of Param and estimation error (MPJPE, lower is better). Our approach achieves competitive performance metrics while maintaining a lightweight model.}
\end{figure}

On the other hand, deterministic methods~\cite{zeng2020srnet,zheng20213d,PSTMO} focus on producing a single definite 3D pose for each image, making them more practical for complex scenarios. However, these models often struggle with the inherent ambiguity in the data, resulting in suboptimal outcomes, especially in complex and challenging scenarios. To conquer these challenges, recent studies employ transformer~\cite{einfalt2023uplift,tang20233d}, graph convolution network (GCN)~\cite{xu2021graph,mehraban2024motionagformer}, and diffusion model~\cite{gong2023diffpose,FinePOSE} to enhance accuracy. Although each of these techniques has its own advantages, none have been integrated into a single hybrid network. Our method addresses this gap by introducing a Hybrid Diffusion-Transformer-GCN (HDiffTG) architecture, which offers an effective solution for 3D human pose estimation, achieving both lightweight design and high accuracy. To the best of our knowledge, we are the first to integrate all these techniques for 3DHPE.

We designed a dual-stream network architecture that integrates Transformer and GCN in a parallel manner to simultaneously extract local graph structures and global sequence relationships, based on which the 3D pose is regressed. To further enhance the model's performance, we combined the Transformer-GCN architecture with a diffusion model for fine-grained optimization of the regressed pose. To achieve efficiency, during the training phase, we directly predict the ground truth 3D pose using noisy, perturbed 3D poses and 2D keypoint conditions. This design significantly reduces computational costs in practical applications and improves prediction accuracy during the sampling phase. Additionally, we interpret the attention mechanism of the Transformer as the discretized form of an underlying partial differential equation, where temporal structures and human topology are discretized into frame-wise inputs. This strategy effectively suppresses the accumulation of highly similar information during the aggregation process,alleviating the issue of over-smoothing caused by the aggregation of highly similar features.

Our contributions are summarized as follows:

\begin{itemize} \item \textbf{Lightweight Hybrid Architecture:} We introduce HDiffTG, a lightweight hybrid Diffusion-Transformer-GCN architecture for high-precision 3D human pose estimation. This design combines Transformer and GCN in a parallel network to capture long-range dependencies and spatiotemporal features, while utilizing a diffusion module for fine-grained optimization of pose predictions.

\item \textbf{Efficient Optimization Strategy:} We propose an efficient objective function optimization strategy combined with diffusion models and design an embedding dimension transformation mechanism at the output layer. This strategy significantly reduces the computational complexity and parameter size of the model by decreasing the number of iterations in the diffusion process, while simultaneously improving the inference speed of 3D human pose estimation.

\item \textbf{Flow Control Mechanism:} We propose a partial differential equation to control the speed of information flow between joints, effectively reducing over-smoothing caused by similar feature aggregation.

\item \textbf{SOTA Performance:} We validate the effectiveness of the proposed model on two benchmark 3D human pose estimation datasets, Human3.6M and MPI-INF-3DHP. Experimental results demonstrate that HDiffTG outperforms existing methods in terms of both accuracy and model efficiency, achieving state-of-the-art performance on the MPI-INF-3DHP dataset. Moreover, the model exhibits exceptional robustness to noisy 2D keypoint inputs, showcasing its significant practical application potential and performance advantages. \end{itemize}

\section{Related Works}
\subsection{Transformer based methods}
In 3D human pose estimation, PoseFormer~\cite{zheng20213d} pioneers the use of Transformer architecture, incorporating both temporal and spatial dimensions to estimate 3D poses from video sequences. PoseFormerV2~\cite{PoseFormerV2} enhances computational efficiency through frequency domain representation and increases robustness to sudden movements in noisy data. METRO~\cite{METRO} introduces an end-to-end Transformer-based network that performs both human pose estimation and mesh reconstruction, effectively integrating 2D features with 3D shape information. MHFormer~\cite{MHFormer} generates multiple hypotheses in the spatial domain and facilitates communication across hypotheses in the temporal domain to synthesize a final pose, addressing self-occlusion and depth ambiguity issues. P-STMO~\cite{PSTMO} employs masked pose modeling to reconstruct 2D poses and reduces errors through a self-supervised pretraining model, easing the capture of spatiotemporal information. MixSTE~\cite{MixSTE} alternates between spatial and temporal Transformer modules in a seq2seq manner, where the spatial module models joint correlations, and the temporal module models motion. HDFormer~\cite{HDFormer} combines self-attention and higher-order attention mechanisms to effectively tackle challenges in complex and highly occluded scenes. STCFormer~\cite{STCFormer} reduces computational complexity by separating correlation learning into spatial and temporal components. Although these methods achieve satisfactory results, the human joints and skeleton information is not well captured.

\subsection{Graph Convolutional Network (GCN)-based methods}
SemGCN~\cite{SemGCN} captures semantic information not explicitly represented in the graph, such as local and global node relationships, by representing 2D and 3D human poses as structured graphs to encode joint relationships in the human skeleton. GnTCN~\cite{GnTCN} introduces GCNs for both human joints and skeletons, using directed graphs and confidence scores from 2D pose estimators to improve pose estimation while modeling skeletal connections, providing information beyond just joints. LCN~\cite{LCN} addresses GCN limitations by assigning dedicated filters to different joints and is trained alongside a 2D pose estimator to handle inaccurate 2D poses. In~\cite{zouZHIMING}, a higher-order GCN is introduced for 3D HPE, enhancing the model's ability to handle complex pose estimation by aggregating node features at various distances through higher-order graph convolutions. Although GCN methods offer high computational efficiency in 3D human pose estimation, they fall short compared to Transformer-based models due to their primary focus on local joints. GLA-GCN~\cite{GLAGCN} presents an adaptive GCN method that uses global representations and a stepwise design to reduce temporal scope, maintaining low memory load while achieving competitive 3D human pose estimation results compared to Transformer-based models. However, the effectiveness of this module in extracting global representations has not yet matched that of attention modules.

\subsection{Diffusion based methods}
 Diffusion model~\cite{ho2020denoising} based methods have recently emerged as a powerful approach for 3D human pose estimation, offering a robust framework for managing uncertainty and generating accurate pose predictions. DiffPose~\cite{holmquist2023diffpose} utilizes a diffusion model to initialize the 3D pose distribution with 2D pose heatmap and depth distributions, constructing a forward diffusion process based on a Gaussian mixture model. GFPose~\cite{GFPose} introduces a time-dependent score network that estimates gradients for each body joint, progressively denoising the perturbed 3D poses to enhance accuracy. D3DP~\cite{shan2023diffusion} proposes a joint-level aggregation strategy that leverages all generated poses to provide a comprehensive estimation, effectively addressing issues related to occlusion and ambiguity. FinePose~\cite{FinePOSE} employs learnable modifiers to achieve multi-granularity control, incorporating coarse and fine-grained human parts and kinematic information to refine the pose estimation. These probabilistic methods usually add t-step noise directly to the original 3D pose during the forward process which is not conducive to learning a clear human pose prior.

We can see that each type of method has its own advantages and disadvantages. Our goal is to integrate the strengths of all these approaches into a single architecture that operates deterministically, generating a single, definitive 3D pose for each image. This approach aims to address practical challenges and reduce ambiguity.

\section{Method}
\begin{figure*}
    \centering
        \includegraphics[width=0.95\linewidth]{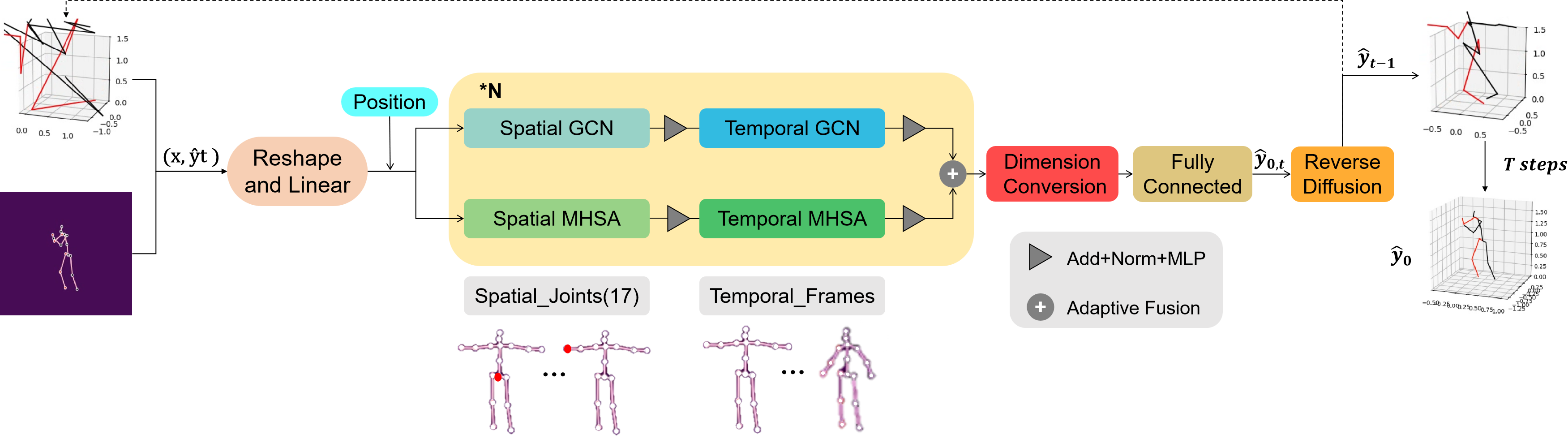}
    \caption {\textbf{The architecture of HDiffTG.} HDiffTG consists of N parallel dual-stream fusion modules combining Transformers and GCNs. The spatial stream processes individual human joints (17 in total), while the temporal stream operates on whole-body poses across frames.} 
    \label{fig:network}
\end{figure*}

The 2D keypoint sequence $\mathbf{X}\in\mathbb{R}^{N\times J\times2}$ consists of \textit{N} frames, each containing \textit{J} keypoints. Our goal is to predict the 3D pose sequence $\mathbf{Y}\in\mathbb{R}^{N\times J\times3}$ for all frames. HDiffTG achieves accurate 3D human pose estimation using a dual-stream network that combines parallel Transformers and GCNs as the backbone, followed by a diffusion model for further refinement. To address the over-smoothing issue caused by numerous iterations in the diffusion model and to reduce the parameter count, we incorporate smoothing techniques based on partial differential equations into the self-attention mechanism and adjust the embedding dimensions in the output head. Additionally, we improve efficiency by significantly reducing the number of iterations required by modifying the objective function. These innovations are integrated into a unified framework, substantially improving the effectiveness of 3DHPE. Fig.~\ref{fig:network} provides an overview of this architecture.

\subsection{Transformer-GCN Dual-Stream Module}

Transformers effectively capture long-range dependencies through self-attention mechanisms, while Graph Convolutional Networks aggregate and transform node features based on the graph's topology, thereby capturing local neighborhood information. Both techniques have unique strengths, and combining them allows for the simultaneous extraction of local graph structures and global sequential relationships, resulting in more efficient feature representations. 

To achieve this goal, we employ a dual-stream structure similar to MotionBert \cite{MotionBERT} which has demonstrated superior performance. However, MotionBert is large in scale and computationally expensive. To address this limitation, we propose incorporating GCNs to leverage their efficiency and the global modeling capacity of Transformers. This enables the design of a more lightweight yet effective framework. We believe that integrating Transformers with GCNs can significantly reduce computational costs while achieving a better balance between local and global feature extraction, thereby fully exploiting the spatiotemporal characteristics of 3D human poses.

In the temporal dimension, the input is rearranged as $F_T\in\mathbb{R}^{BJ\times T\times d}$,where each input token represents a human body joint. In the spatial dimension, the input is rearranged as $F_S\in\mathbb{R}^{BT\times J\times d}$, where each input token corresponds to a frame in the pose sequence.

Within the Transformer framework, we employ the classical Multi-Head Self-Attention (MHSA) mechanism. The spatial self-attention module is designed to model the relationships between human body joints within the same time step, capturing their global dependencies. Meanwhile, the temporal self-attention module enhances the modeling of human dynamic movements by effectively capturing the changes in joint motion across the time sequence.

Unlike Transformers, which focus on aggregating global information, GCNs place greater emphasis on capturing local spatial and temporal relationships within skeletal sequences. Through ablation analysis in Section~\ref{tab:Ablation Studies}, it is demonstrated that the parallel dual-stream design incorporating GCN significantly improves the model's adaptability and representational capacity when handling complex information.

In the GCN architecture, the construction of the adjacency matrix varies based on the token inputs from the spatial and temporal dimensions. In the spatial GCN, the adjacency matrix represents the topological structure of the human body to capture spatial relationships within the skeletal sequence. In contrast, in the temporal GCN, the adjacency matrix is dynamically generated based on temporal similarity between node features. The temporal similarity is calculated as follows:
\begin{equation}
    \mathrm{Similar}(\mathbf{F}_{t_i},\mathbf{F}_{t_j})=\mathbf{F}_{t_i}\cdot\mathbf{F}_{t_j}^T,
\end{equation}
where $\mathbf{F}_{t_{i}}$ and $\mathbf{F}_{t_j}$ denote the feature vectors at time steps \textit{i} and \textit{j}, respectively. 

Based on this similarity matrix, the KNN approach is further applied to select the K most similar neighboring nodes for each time step. The temporal adjacency matrix effectively captures dynamic relationships between nodes in the temporal dimension, while leveraging local similarity constraints to significantly enhance the precision of temporal modeling.

\textbf{Adaptive Fusion.} We use adaptive fusion to aggregate features extracted by the Transformer and GCN streams. The fusion weights dynamically balance according to the spatiotemporal characteristics of the input which is defined as:
\begin{equation}
    F^{(i)}=\alpha_{T}^i\circ F_{T}^{(i-1)}+\alpha_{G}^i\circ F_{G}^{(i-1)},
\end{equation} where $\circ$ represents the element-wise multiplication, $F^{(i)}$ represents the feature embedding at depth \textit{i}, $F_{T}^{(i-1)}$ and $F_{G}^{(i-1)}$ represents the features extracted at depth \textit{(i-1)} from the Transformer stream and the GCN stream, respectively. The adaptive fusion weights $\alpha_{T}$ and $\alpha_{G}$ are obtained through the following equations:\begin{equation}
    \alpha_{T}^i,\alpha_{G}^i=\mathrm{softmax}(W(F_{T}^{(i-1)} , F_{G}^{(i-1)}))),
\end{equation} where \textit{W} is a learnable linear transformation, and [,] denotes concatenation.

\subsection{Diffusion Based Refinement}

In real-world scenarios, 2D keypoint inputs often suffer from occlusion issues, and critical depth information is frequently lost during the conversion from 2D coordinates to 3D. Although the parallel dual-stream architecture of Transformers and GCNs is capable of capturing both global and local spatiotemporal relationships, it remains insufficiently robust when faced with occlusion and missing information. Therefore, we introduce a diffusion framework to enhance the model's predictive accuracy and robustness in complex environments.

We define the diffusion process as \textit{q}, where noise is progressively added by sampling time steps $\mathrm{t}\sim\mathrm{U}(0,\mathrm{T_{M}})$ from a uniform distribution, with \textit{$T_{M}$} representing the maximum time step. During the diffusion process, the initial random noisy pose $y_{t}$ is gradually denoised to generate the target 3D pose sequence $\hat{y}_0$. In the reverse diffusion process at step \textit{t}, the noisy 3D sequence $\hat{y}_t$ is progressively updated based on the predictions of the backbone model $f_\theta$.The goal of the backbone model is to predict the denoised 3D sequence $\hat{y}_{0,t}$ at the current time step, given the 2D keypoint sequence \textit{x}. This process can be formulated as: $\hat{y}_{0,t}=f_\theta(x,\hat{y}_t,t).$

We define the reverse diffusion process using a predefined reverse diffusion function, where the predicted denoised 3D pose $\hat{y}_{0,t}$ is combined with the current noisy state $\hat{y}_{T}$ to obtain the noisy 3D sequence of the previous time step, $\hat{y}_{t-1}$:
\begin{equation}
    \hat{y}_{t-1}=\mu_t+\sigma_t\cdot z,
\end{equation}
where ${z}$ represents random noise, ${z}\sim\mathcal{N}(0,I)$. $\mu_t$ and $\sigma_t$ represent the mean and standard deviation of the posterior distribution, respectively. These are defined as follows:
\begin{equation}
\mu_{t}=\frac{\sqrt{\bar{\alpha}_{t-1}}\beta_{t}}{1-\bar{\alpha}_{t}}\hat{y}_{0,t}+\frac{\sqrt{\alpha_{t}}(1-\bar{\alpha}_{t-1})}{1-\bar{\alpha}_{t}}\hat{y}_{t}, \\
\end{equation}
\begin{equation}
\sigma_t^2=\frac{\beta_t(1-\bar{\alpha}_{t-1})}{1-\bar{\alpha}_t},
\end{equation}
By iteratively applying this process, the random noise sequence $\hat{y}_{T}$ is progressively transformed into the target denoised 3D pose sequence $\hat{y}_{0}$.

However, the standard reverse diffusion process typically requires \textit{T}=1000 sampling steps, which significantly increases computational costs. To address this issue, we adopt the improved DDIM (Denoising Diffusion Implicit Models) method, which accelerates the reverse diffusion process by modifying the update rule. The update formula for time step $\tau_{i}$ is given as:
\begin{equation}
    \hat{y}_{\tau_{i-1}}=w_1\cdot\hat{y}_{0,\tau_i}+w_2\cdot\hat{y}_{\tau_i},
\end{equation}
\begin{equation}
    w_{1}=\bar{\alpha}_{\tau_{i-1}}-\frac{\sqrt{1-\bar{\alpha}_{\tau_{i-1}}}\cdot\bar{\alpha}_{\tau_{i}}}{\sqrt{1-\bar{\alpha}\tau_{i}}}\quad w_{2}=\frac{\sqrt{1-\bar{\alpha}_{\tau_{i-1}}}}{\sqrt{1-\bar{\alpha}_{\tau_{i}}}}
\end{equation}
Finally, the model obtains the denoised 3D pose sequence via $\hat{y}_{0}=\hat{y}_{0,\tau_1}$.

To further reduce the number of sampling steps, we refine the objective of the backbone model to directly predict the clean 3D pose $\hat{y}_{0,t}$ at step \textit{t}.Under this framework, the prediction of Gaussian noise can be reformulated as:
\begin{equation}
    \hat{\epsilon}_t=\frac{\hat{y}_t-\bar{\alpha}_t\hat{y}_{0,t}}{\sqrt{1-\bar{\alpha}_t}}
\end{equation}
With this adjustment, the diffusion model is capable of generating high-quality 3D poses while significantly reducing the number of sampling steps, thereby lowering computational complexity.

\textbf{Over-Smoothing Handling.} The core idea of the diffusion model is to iteratively diffuse node features, allowing information to be shared between adjacent nodes. However, when the number of iterations becomes too large, the high similarity between the same joints in consecutive frames and between adjacent joints in the same frame causes the node features to become increasingly uniform over time. This leads to a loss of differentiation between nodes, resulting in the phenomenon known as "over-smoothing."

To address this issue, We propose a processing method based on partial differential equations (PDE), where the attention mechanism of the Transformer is interpreted as the discretized form of an underlying PDE. By doing so, we discretize the temporal structures and human topology into frame-wise inputs and design a corresponding diffusion operator to effectively suppress the accumulation of highly similar information during the aggregation process, thereby preventing over-smoothing. The proposed partial differential equation is defined as:
\begin{equation}
    \hat{y}_t=\hat{y}_0+\int_0^t(A(\hat{y}_\tau)-I)\cdot\hat{y}_\tau d\tau
\end{equation}
where the learned node embeddings $\hat{y}$ is defined as $\hat{y}=\phi(\hat{y}_0)$, and satisfy the following equation:
\begin{equation}
    \hat{y}_0=y_T-\int_0^T\frac{\partial\hat{y}_t}{\partial t}dt
\end{equation}
where $A(\hat{y}_{t})$ represents the adjacency matrix at time step \textit{t}, and \textit{I} is the identity matrix. This partial differential equation dynamically adjusts the relationships between nodes via the adjacency matrix, enabling the diffusion process to model temporal and spatial dependencies effectively.

\section{Experiments}
\subsection{Datasets and Metrics}
We evaluated the proposed HDiffTG by comparing to the state-of-the-art 3D human pose estimation methods on the Human3.6M~\cite{Human36M} and MPI-INF-3DHP datasets~\cite{3DHP}.

\textbf{MPI-INF-3DHP} is a large-scale dataset containing over 1.4 million frames captured using 14 cameras in both indoor and outdoor environments. The dataset includes diverse actions performed by 8 actors, including complex movements. We evaluated this dataset using the Percentage of Correct Keypoints (PCK), Area Under the Curve (AUC) (within a 150mm range), and the Mean Per Joint Position Error (MPJPE). MPJPE calculates the average Euclidean distance (in millimeters) between the predicted 3D joint coordinates and the ground truth. The dataset contains over 2,000 videos with 13 annotated keypoints in outdoor scenes, making it highly suitable for both 2D and 3D pose estimation. Its markerless multi-camera system enriches data for both foreground and background scenes, enhancing the model's generalization capability, particularly under occlusions and complex scenarios.

\textbf{Human3.6M} is a widely-used large-scale dataset for 3D human pose estimation, collected in an indoor environment with four cameras at different angles, providing a total of 3.6 million accurate human poses. We evaluated methods on this dataset using MPJPE and Procrustes-MPJPE (P-MPJPE). P-MPJPE is calculated by first rigidly aligning the predictions to the ground truth and then computing the adjusted MPJPE.

\subsection{Experimental Setup}
Our HDiffTG is implemented using the PyTorch and trained on three GeForce RTX 3090 GPUs. We trained the model from scratch in an end-to-end manner, using the Adam~\cite{Adam} optimizer with a weight decay of 0.1. The initial learning rate is set to 0.0005, and after each epoch, the learning rate is multiplied by a decay factor of 0.99. The dropout rate was set to 0.1, and the forward diffusion steps \textit{T} were set to 1000. For a fair comparison, we apply the same horizontal flipping augmentation as used in SemGCN~\cite{SemGCN}. On the Human3.6M dataset, we evaluated the methods using 2D keypoints detected by CPN~\cite{CPN}. For the MPI-INF-3DHP dataset, we followed the protocol used in PoseFormer~\cite{zheng20213d} where the ground truth 2D keypoints of 17 joints as input. We validated our network on the test set to ensure consistent evaluation.

\subsection{Results and Analysis}
\begin{figure}
    \centering
    \includegraphics[width=0.92\linewidth]{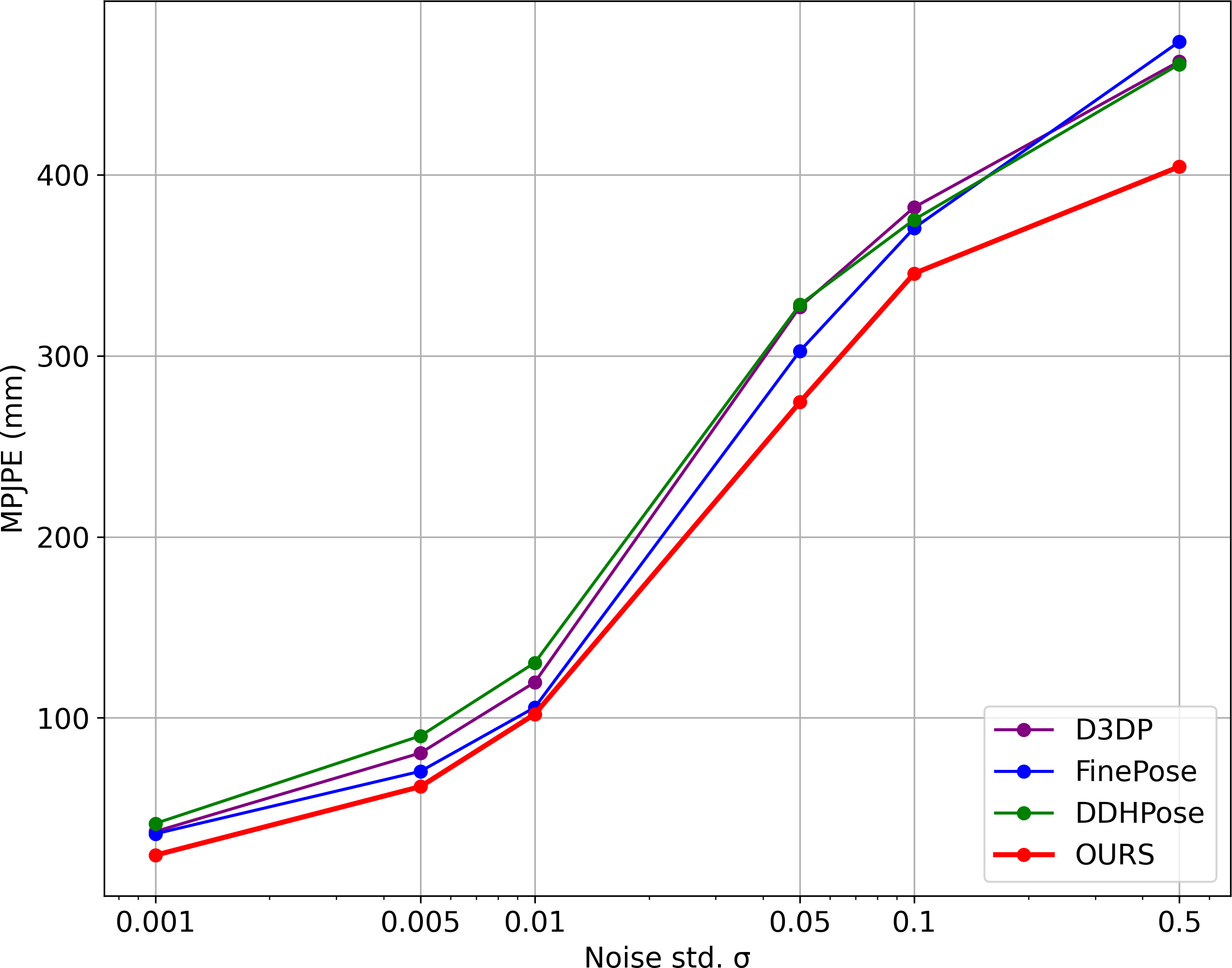}
    \caption{The performance variations of HDiffTG, D3DP, FinePose, and DDHPose on the MPI-INF-3DHP dataset when Gaussian noise with a mean of 0 and standard deviation $(\sigma)$  is added are analyzed.}
    \label{fig:enter-label}
\end{figure}

\begin{figure*}[ht]
    \centering
        \includegraphics[width=0.90\linewidth]{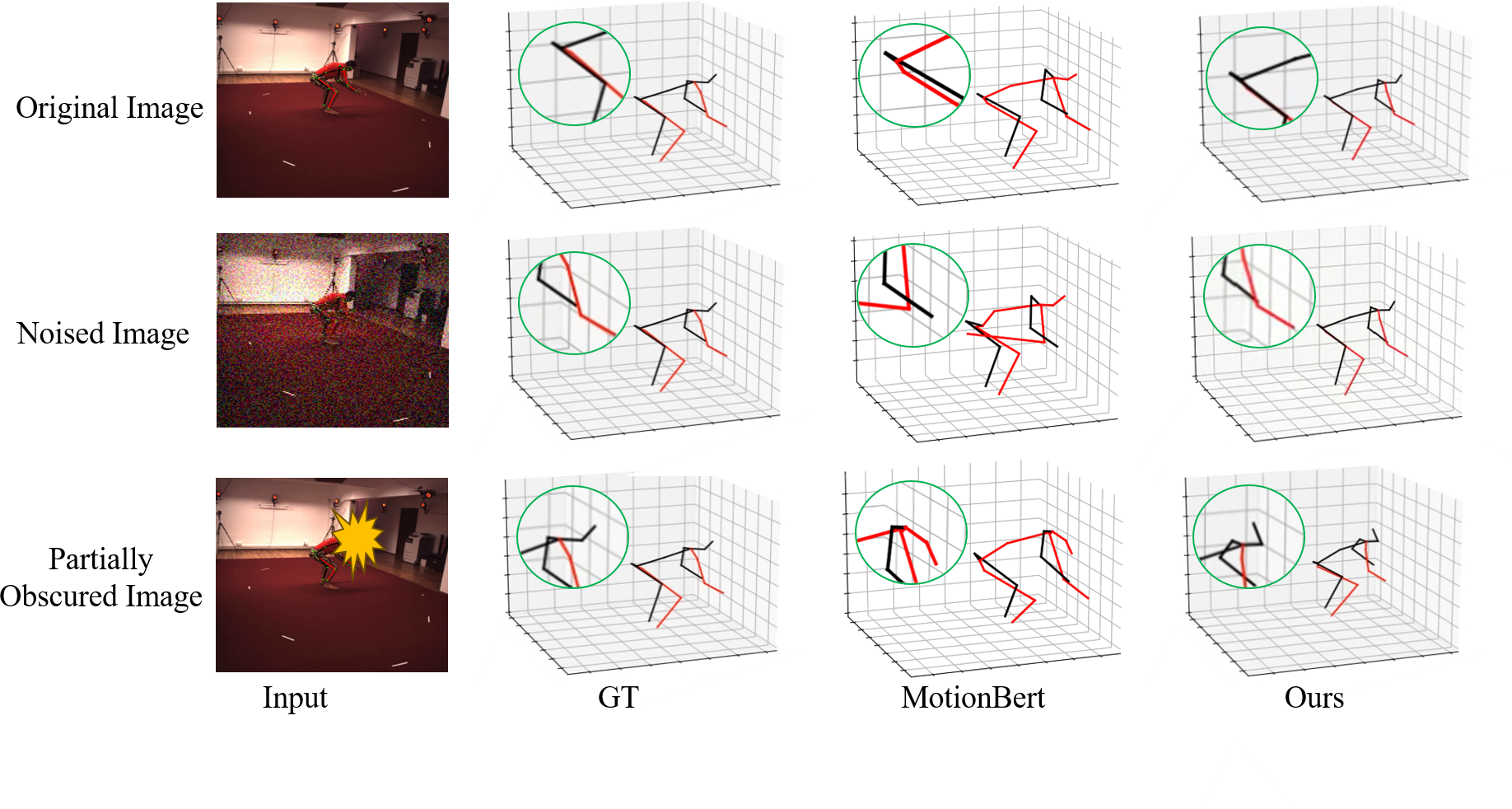}
        \vspace{-30pt}
    \caption{Qualitative comparison of our HDiffTG with the state of the art 3D pose estimation approach, MotionBert~\cite{MotionBERT} on Human3.6M under noise and joint occlusion.}
    \label{fig:compare}
\end{figure*}

\begin{table}[ht]\small
      \centering
        \caption{Quantitative comparisons on MPI-INF-3DHP. The best and second-best scores are in bold and underlined, respectively.}
        \begin{tabular}{l|cccc}
          \hline
          Method  & PCK$\uparrow$ & AUC$\uparrow$ & MPJPE(mm)$\downarrow$ \\
          \hline
          P-STMO~\cite{PSTMO} ECCV’22& 97.9 & 75.8 & 32.2 \\
          MHFormer~\cite{MHFormer} CVPR’22  & 93.8 & 63.3 & 58.0 \\
          MixSTE~\cite{MixSTE} CVPR’22 & 96.9 & 75.8 & 35.4 \\
          HDFormer~\cite{HDFormer} IJCAI’23 & \underline{98.6} & 72.9 & 37.2 \\ 
          Diffpose~\cite{gong2023diffpose} CVPR’23 & 98.0 & 75.9 & 29.1 \\
          STCFormer~\cite{STCFormer} CVPR’23 & \underline{98.6} & \underline{83.9} & \underline{23.1} \\
          PoseFormerV2~\cite{PoseFormerV2} CVPR’23 & 97.9 & 78.8 & 27.8 \\
          GLA-GCN~\cite{GLAGCN} ICCV’23 & 98.5 & 79.1 & 27.8 \\
          D3DP~\cite{shan2023diffusion} ICCV’23 & 97.7 & 77.8 & 30.2 \\
        DDHPose~\cite{DDHPose} AAAI’24 & 98.5 & 78.1 & 29.2 \\
          FinePose~\cite{FinePOSE} CVPR’24 & \textbf{98.7} & 80.0 & 26.2 \\
          Ours & \textbf{98.7} &\textbf{ 85.2} & \textbf{18.2} \\
          \hline
        \end{tabular}%
    \label{tab:MPI-INF-3DHP-comparison}
    \end{table}

\begin{table}[ht]\small
      \centering
        \caption{Quantitative comparisons on Human3.6M dataset. The best and second-best scores are in bold and underlined, respectively.}
        \begin{tabular}{l|cccc}
          \hline
          Method  & Param & MPJPE(mm)$\downarrow$ & P-MPJPE(mm)$\downarrow$ \\
          \hline
          P-STMO~\cite{PSTMO}& 7.0 M & 42.8 & 34.4 \\
          MHFormer~\cite{MHFormer}  & 30.9 M & 42.9 & 34.4 \\
          MixSTE~\cite{MixSTE} & 33.6 M & 40.9 & 32.6 \\
          HDFormer~\cite{HDFormer} & 98.7 M & 42.6 & 33.1 \\ 
          PoseFormerV2~\cite{PoseFormerV2} & 14.3 M & 45.2 & 35.6 \\
          GFPose~\cite{GFPose} & 98.7 M & 45.1 & 38.4 \\
          STCFormer~\cite{STCFormer} & 18.9 M & 40.5 & 31.8 \\
          GLA-GCN~\cite{GLAGCN} & 1.3 M & 44.4 & 34.8 \\
          D3DP~\cite{shan2023diffusion}& 34.9 M & 40.1 & 31.6 \\
    
          MotionBert~\cite{MotionBERT} & 42.5 M & \textbf{39.2} & 32.9 \\
          DDHPose~\cite{DDHPose} & 38.5 M & \underline{39.6} & \textbf{31.2} \\
          Ours & 7.5 M &39.9 & \underline{31.4} \\
          \hline
        \end{tabular}%
    \label{tab:h36m_results}
    \end{table}    

    \begin{table}[t]
 \centering
  \caption{\textit{d}: The embedding dimension of the input.\textit{$d^{\prime}$}:The embedding dimension before the regression head.Bold indicates our HDiffTG model.}
 \begin{tabular}{@{}lccc@{}}
   \hline
   $\mathrm{d-d}^{\prime}$ & Params (M) & MPJPE(mm)$\downarrow$ & P-MPJPE(mm)$\downarrow$ \\
   \hline
   \textbf{128-512} & 7.50 &\textbf{39.9} & \textbf{31.4}\\
   128-256 & 7.48 & 44.9 & 36.9\\
   128-1024 & 7.52 & 45.7 & 37.7\\
   64-512 & 1.9 & 41.3 & 34.4 \\
   256-512 & 29.7 & 40.3 & 32.6 \\
   512-512 & 118.1 & 40.8 & 33.2\\    
   \hline
 \end{tabular}
 \label{tab:dimension}
\end{table}

\begin{table}[ht]\small
      \centering
        \caption{Comparison of different diffusion frameworks used in 3D human pose estimation tasks in terms of model parameters size and frames per second (FPS).}
        \begin{tabular}{l|cccc}
          \hline
          Method  & Params(M) & frame/s$\uparrow$ & MPJPE(mm)$\downarrow$ \\
          \hline
          Diffpose~\cite{gong2023diffpose} & 30.9 & 376  & 29.1 \\
          D3DP~\cite{shan2023diffusion} & 34.8 & 70  & 30.2 \\
        DDHPose~\cite{DDHPose} & 38.5 & 1634 & 29.2 \\
          FinePose~\cite{FinePOSE} & 200.6 & 7  & 26.2 \\
          Ours & \textbf{7.5} & \textbf{2922} & \textbf{18.2} \\
          \hline
        \end{tabular}%
    \label{tab:comparison}
    \end{table}

Studies~\cite{MHFormer} have shown that models trained and tested on the MPI-INF-3DHP dataset demonstrate superior performance in handling occlusion challenges. This is because the dataset includes human pose data captured in both indoor and outdoor environments, covering a wide range of lighting conditions and backgrounds. In contrast, the Human3.6M dataset primarily focuses on indoor laboratory settings, which feature more uniform scenes and lighting. The research results indicate that the diversity and challenging conditions of the MPI-INF-3DHP dataset encourage better model performance in occlusion scenarios. Our model’s outstanding performance on this dataset (Table~\ref{tab:MPI-INF-3DHP-comparison}), significantly surpassing other existing 3D human pose estimation methods and achieving state-of-the-art standards, further demonstrates the robustness of HDiffTG against disturbances.

The results on the Human3.6M dataset are shown in Table~\ref{tab:h36m_results}. Since HDiffTG is a deterministic method, we compared it with the latest deterministic methods that use the same number of frames (243 frames) and can measure the specific error of each pose. The comparison only included models that were not pre-trained on additional data. As shown, HDiffTG achieved an MPJPE of 39.8mm and a P-MPJPE of 31.2mm, surpassing most of the latest methods.

To evaluate the robustness of HDiffTG in more challenging scenarios, we designed an artificial Gaussian noise. This noise has a mean of 0, and the standard deviation $(\sigma)$ is set to 0.001, 0.005, 0.01, 0.05, 0.1, and 0.5, respectively. It is directly added to the $(x, y)$ coordinates of the 2D keypoints to simulate localization errors that may occur in real 2D detectors. The standard deviation $(\sigma)$ of the Gaussian noise represents the magnitude of the perturbation on the keypoint coordinates, with its unit being consistent with that of the 2D keypoint coordinates, measured in pixels. The experimental results, as shown in Fig.~\ref{fig:enter-label}, indicate that as the noise level increases, the performance of all methods on the MPI-INF-3DHP dataset declines significantly. However, compared to baseline methods, HDiffTG still demonstrates strong robustness in different noisy scenarios, particularly under high noise levels, where its performance degradation is significantly smaller than that of other methods.

Fig.~\ref{fig:compare} illustrates the qualitative comparison with the current state-of-the-art method, MotionBert~\cite{MotionBERT}. In more challenging scenarios, HDiffTG exhibits superior performance, showcasing its robustness under difficult conditions.

Most 3D human pose estimation models based on diffusion frameworks have inherent advantages in handling depth ambiguity and 2D pose estimation errors. However, these methods are often accompanied by high computational complexity. In contrast, the HDiffTG model reduces the number of iterations during the diffusion process by optimizing the objective function and transforming the embedding dimensions at the output layer, thereby significantly decreasing the model size and testing time (as shown in Table~\ref{tab:dimension} and Table~\ref{tab:comparison}). Experimental results demonstrate that among diffusion models focused on the 3D human pose estimation task, HDiffTG has the smallest number of parameters and achieves the highest FPS (frames per second) under the same frame count (243 frames). Notably, this FPS refers to the speed of 2D-to-3D conversion rather than the speed of the entire 3D human pose estimation process. Through the aforementioned design, HDiffTG not only simplifies the model's complexity but also achieves significant performance improvements, providing an efficient solution for applying diffusion models in the field of 3D human pose estimation.

\subsection{Ablation Studies}\label{tab:Ablation Studies}

We performed a series of ablation studies on the MPI-INF-3DHP dataset to evaluate the effectiveness of the different components within our hybrid architecture.

First, we test the effect of the PDE. The results are shown in Table~\ref{tab:PDE}. All models use the same number of forward diffusion steps, iterating 1000 times, but with different numbers of layers. The experimental results demonstrate that the PDE module significantly improves the performance of the HDiffTG on the MPI-INF-3DHP dataset, particularly when the number of layers is smaller. However, as the model depth increases, the positive impact of the PDE module may become limited, and the model may face the risk of overfitting. In contrast, models without the PDE module exhibit higher MPJPE across all layers, indicating that the PDE module plays a critical role in enhancing the model’s representation capability and optimizing stability.

\begin{table}[t]\small
  \centering
    \caption{Comparison of the impact of applying PDE smoothing versus omitting it across various  layers on the MPI-INF-3DHP dataset. \textbf{Bold} indicates the default setting for our HDiffTG.}
  \begin{tabular}{l|ccccc}
    \hline
    layers & 4 & 6 & 8 & 10 & 12 \\
    \hline
    w PDE ~ & 23.76 & 23.33 & \textbf{18.20} & 21.52 & 24.56 \\
    w/o PDE ~ & 54.11 & 51.65 & 48.98 & 49.25 & 33.26 \\ 
    \hline
  \end{tabular}

  \label{tab:PDE}
\end{table}

\begin{table}[t]\small
  \centering
    \caption{Comparison of different integration methods of GCN and Transformer on the MPI-INF-3DHP dataset. \textbf{Bold} indicates our HDiffTG deault setting.}
  \begin{tabular}{@{}lc@{}}
    \hline
    Method & MPJPE (mm) $\downarrow$ \\
    \hline
    GCN only ~ & 50.1 \\
    Transformer only ~ & 22.2 \\
    GCN $\rightarrow $ Transformer (Sequential) ~ & 19.9 \\
    Transformer $\rightarrow $ GCN (Sequential) ~ & 19.6 \\  
    GCN - Transformer (Parallel) ~ & \textbf{18.2} \\  
    \hline
  \end{tabular}
  \label{tab:module}
\end{table}

To validate the effectiveness of the Transformer-GCN module, we present the experimental results for different module configurations in Table~\ref{tab:module}. When using only the GCN module, the MPJPE error is 50.1mm, highlighting some limitations in accurately capturing the 3D sequence structure. In contrast, the hybrid approach that combines both GCN and Transformer modules significantly improves performance, reducing the MPJPE error by 3.6mm compared to using the Transformer alone. Furthermore, the results indicate that the parallel integration of these two modules is slightly more effective than sequential fusion.

\section{Conclusion}
In this paper, we introduced a novel Hybrid Diffusion-Transformer-GCN (HDiffTG) architecture for 3D Human Pose Estimation (3DHPE) from monocular visual data. By integrating the strengths of Diffusion, Transformer and GCN, our approach effectively addresses the challenges posed by depth ambiguities, occlusions, and complex human body dynamics. The dual-stream network integrates Transformer and GCN to simultaneously extract local graph structures and global sequence relationships, while the diffusion module further refines the pose estimation. Our lightweight design reduces model parameters, resulting in the most efficient design among similar diffusion models. Experimental results on the Human3.6M and MPI-INF-3DHP datasets confirm that our HDiffTG outperforms existing methods in both accuracy and robustness. This highlights the significant practical value and potential for real-world applications of our approach. 

\section{Acknowledgment}
This research is supported by the Singapore Ministry of Education (MOE) Academic Research Fund (AcRF) Tier 1 grant (22-SIS-SMU-093), Ningbo 2025 Science \& Technology Innovation Major Project (No. 2022Z072), and Fundo para o Desenvolvimento das Ciências e da Tecnologia of Macau (FDCT) with Reference No. 0067/2023/AFJ, No. 0117/2024/RIB2.

{
    \small
    \bibliographystyle{unsrt}
    \bibliography{HDiffTG-CameraReady}
}

\end{document}